\documentclass[runningheads]{llncs}

\usepackage[final,year=2024,ID=5889]{eccv}

\usepackage{eccvabbrv}
\usepackage{graphicx}  
\usepackage{float}  
\usepackage{graphicx}
\usepackage{booktabs}
\usepackage{diagbox}
\usepackage{bm}
\usepackage{multirow}
\usepackage[accsupp]{axessibility}  
\usepackage{threeparttable}
\usepackage{algorithm}  
\usepackage{algpseudocode}  
\usepackage{amsmath}

\usepackage[pagebackref,breaklinks,colorlinks,citecolor=eccvblue]{hyperref}

\usepackage{orcidlink}
\usepackage{caption}
\usepackage{graphicx}
\usepackage{float} 
\usepackage{subcaption}

\begin{document}
\title{Environmental Matching Attack Against Unmanned Aerial Vehicles Object Detection} 

\titlerunning{Abbreviated paper title}

\author{Dehong Kong\inst{1} \and
Siyuan Liang\inst{2} \and
Wenqi Ren\inst{3}}

\authorrunning{F.~Author et al.}

\institute{Shenzhen Campus of Sun Yat-sen University, National University of Singapore \and
Sun Yat-Sen University
}

\maketitle

\begin{abstract}

Object detection techniques for Unmanned Aerial Vehicles (UAVs) rely on Deep Neural Networks (DNNs), which are vulnerable to adversarial attacks. Nonetheless, adversarial patches generated by existing algorithms in the UAV domain pay very little attention to the naturalness of adversarial patches. Moreover, imposing constraints directly on adversarial patches makes it difficult to generate patches that appear natural to the human eye while ensuring a high attack success rate. Inspired by the naturalness of a patch's color when it matches its surroundings, we propose a new method named Environmental Matching Attack (EMA) to address the issue of optimizing the adversarial patch under the constraints of color. To the best of our knowledge, this paper is the first to consider natural patches in the domain of UAVs. 
The EMA method exploits strong prior knowledge of a pre-trained stable diffusion to guide the optimization direction of the adversarial patch, where the text guidance can restrict the color of the patch. 
To better match the environment, the contrast and brightness of the patch are appropriately adjusted. Instead of optimizing the adversarial patch itself, we optimize an adversarial perturbation patch which initializes to zero so that the model can better trade off attacking performance and naturalness.
Experiments conducted on the DroneVehicle and Carpk datasets have shown that our work can reach nearly the same attack performance in the digital attack (no greater than 2$\%$ in mAP), surpass the baseline method in the physical specific scenarios, and exhibit a significant advantage in terms of naturalness in visualization and color difference with the environment.

\end{abstract}

\section{Introduction}
Unmanned Aerial Vehicles (UAVs), commonly known as drones, have undergone a remarkable transformation from niche military tools to ubiquitous assets in a wide range of civilian and commercial sectors. Their applications span from aerial photography, agriculture, and environmental monitoring to emergency response, infrastructure inspection, and logistics. Object detection technology is an important means for UAVs to complete tasks. Similar to most general object detection, UAV object detection also relies on Deep Neural Network (DNN) models. The application of DNNs in the field of drones is becoming increasingly widespread, providing strong technical support for the intelligent development of drones. Drones combined with deep learning algorithms can complete various tasks such as drone swarm coordination~\cite{chung2018survey}, fault diagnosis~\cite{liang2022data}, path planning~\cite{kuang2020real}, and other aspects, further expanding the application scenarios and functions of drones. 

In recent years, DNNs have been susceptible to adversarial attacks~\cite{xie2017adversarial}. In practical applications, attackers can significantly bias the DNN model towards detection evasion with minimal input perturbations. A common attack practice relies on the application of adversarial patches, which has indeed achieved good attack performance. There has been some work~\cite{du2022physical,hartnett2022empirical,lian2022benchmarking,lu2021scale,shrestha2023towards} on transferring the adversarial patch attacks on general object detectors to the UAV domain with certain effects, but it needs to change the pixel values of the patches to guarantee attacking performance, which results in conspicuous patterns that noticed by people easily. Although these patches are optimized primarily for attack performance, achieving a well-balanced solution for both attack performance and naturalness is still challenging.

Some studies~\cite{huang2020universal,tan2021legitimate} have demonstrated that applying appropriate constraints through generative models or other methods during the patch generation process can preserve sufficient semantic information, making the generated adversarial patches appear natural to the human eye. However, these works focus on generating adversarial patches that resemble natural objects, without considering the impact of the environment. 

We notice that adversarial patches matching the environmental color can effectively divert human attention and exhibit a high degree of naturalness. Therefore, environmental naturalness should be considered to examine the naturalness of adversarial patches for adversarial UAV object detectors. To our knowledge, no current research addresses these two issues of adversarial attacks in UAV object detection: the lack of environmental naturalness and the absence of methods to constrain patch colors. In this work, we first experiment with whether the work on naturalness in general adversarial object detection can be directly transferred to the UAV domain. Our experimental results indicate that the direct application of these methods fails to yield results as favorable as those observed in the general domain. We attribute the reasons as follows. UAV imagery is subject to distinct influencing factors, including the drone's altitude, camera specifications, and the surrounding environmental conditions. Owing to these specific attributes, the task of crafting naturally deceptive adversarial patches for UAV images presents significant challenges.


Therefore, we propose the Environmental Matching Attack (EMA) to generate adversarial patches against UAV object detectors and experimentally verify that our method achieves a good trade-off between attack performance and naturalness. The essence of the EMA method is to control the color of the adversarial patch to match the environment while relaxing the constraints on texture. To address the issues of color constraints and environmental matching, we exploit text-guided diffusion models where prompt can conveniently limit the optimization range of the patch during training. Scene Matching technique adds contrast and brightness to get close to the physical environment and applies affine transformations to the patch. Experiments have found that in addressing the optimization problem of limiting the patch, direct optimization of the patch is no longer a good choice. To solve the training issue, we choose a tensor initialized to zero of the same size as the patch as the training parameter, ensuring that there will not be too significant changes in the adversarial patch.

In summary, the main contributions are summarized as follows:
\begin{itemize}
\item[$\bullet$] We are the first to research natural patches in the domain of UAVs and focus more on the impact of the environment on the naturalness of adversarial patches. Our method for UAVs implements a dual adversarial attack, which targets the object detector and the human eyes, respectively.
\item[$\bullet$] We propose a novel method named Environmental Matching Attack(EMA) to optimize the adversarial patch under the constraints of color. Text-guided Stable Diffusion and scene matching are adopted to achieve consistency in patch and environment colors. In training, we use the way of adding perturbations instead of directly optimizing the patch, making the training process more controllable and effective.
\item[$\bullet$]We experimentally evaluate the attacking performance and naturalness of our patches. In the digital domain, our natural patches achieve attack performance close to the baseline method with mAP not exceeding 2 and significantly surpass the patches that consider naturalness. In the specific physical environment, our patches have better performance than the baseline. 
\end{itemize}

\section{Related Work}
\subsection{Adversarial Patch}
Adversarial patch attack aims to generate image-independent patches to fool the detector. It can be mainly divided into iterative-based and generative-based methods.

\textbf{Iterative-based Methods}. Brown \etal~\cite{brown2017adversarial} present a method to create universal, robust, targeted adversarial image patches in the real world.
DPatch~\cite{liu2018dpatch} generates a black-box adversarial patch attack for mainstream object detectors by randomly sampling adversarial patch locations and simultaneously attacking the regression module and classification module of the detection head.
Based on DPatch, Lee \etal~\cite{lee2019physical} use the PGD~\cite{madry2017towards} optimization method as a prototype to generate a more aggressive attack method by randomly sampling patch angle and scale changes.
Pavlitskaya \etal~\cite{pavlitskaya2022suppress} also reveal that the adversarial patch scale is proportional to the attack success rate.
Thys et al.~\cite{thys2019fooling} introduce an adversarial patch attack designed to attack person detection in the physical domain. The authors applied various transformations to the patch, such as enhancing printability and reducing pixel variation, to create more 'printable' patches. This work has had a significant influence on the field of UAV adversarial attacks.
Saha \etal~\cite{saha2020role} analyze the attack principle of adversarial patches that do not overlap with the target and propose to use contextual reasoning to fool the detector.
To reduce patch visibility and enhance the attacking ability of the adversarial patch, a large number of works have made a lot of efforts to generate various patches.Specifically, they include adversarial semantic contours that target instance boundaries~\cite{zhang2021adversarial}, adversarial patch groups at multiple locations~\cite{zhao2020object, zhu2021you}, patch-based sparse adversarial attacks~\cite{bao2020sparse}, diffuse patches of asteroid-shaped or grid-shape~\cite{wu2010dpattack}, deformable patch~\cite{chen2022shape} and the translucent patch~\cite{zolfi2021translucent}.

\textbf{Generative-based Methods}.
Attacking ability is not the only goal we pursue. The mainstream method to generate an adversarial patch currently is iterative-based which can optimize for the patch to attack the detector without any constraints, while the patch will be generated in an unpredictable direction. 
To address this problem, generative-based methods are considered to trade off Naturalness for attack performance. 
PS-GAN~\cite{liu2022efficient} proposes a perceptual-sensitive generative adversarial network that treats the patch generation as a patch-to-patch translation via an adversarial process, feeding any types of seed patch and outputting the similar adversarial patch with high perceptual correlation with the attacked image. Pavlitskaya \etal~\cite{pavlitskaya2022feasibility} have shown that using a pre-trained GAN helps to gain realistic-looking patches while preserving the performance similar to conventional adversarial patches. 
Hu \etal.~\cite{hu2021naturalistic} presents a technique for creating physical adversarial patches for object detectors by utilizing the image manifold learned by a pre-trained GAN on real-world images.
Diff-PGD~\cite{xue2024diffusion} utilizes a diffusion model-guided gradient to ensure that adversarial samples stay within the vicinity of the original data distribution while preserving their adversarial potency.

\subsection{UAV Adversarial Attack}
Many works have transferred the methods of general adversarial object detectors to the UAV domain, and they have achieved relatively good performance in terms of attacking performance. However, the research on UAV adversarial attacks is still in its infancy. Hartnett \etal~\cite{hartnett2022empirical} considers adversarial patches in real scenarios from the perspective of overhead images taken by aerial or satellite cameras. 
The AP-PA~\cite{lian2022benchmarking} proposes a new framework to generate adversarial patches that are adaptive for the varying scales of UAV imagery. The authors establish one of the first comprehensive, coherent, and rigorous benchmarks to evaluate the attack efficacy of adversarial patches on aerial detection tasks.
Similarly, Patch-Noobj~\cite{lu2021scale} adaptively scales the width and height of the adversarial patch according to the size of the object.
Andrew Du \etal~\cite{du2022physical} implements an adversarial patch attack for cars on aerial imagery, which is unable to fully depict the specific situation of UAV imagery. Instead, Shrestha \etal~\cite{shrestha2023towards} takes full account of UAV imagery like the VisDrone dataset, models the printed colors Gaussian noise, and adjusts the patch to an easier-to-be-printed format. However, the above methods only improve attacking ability and do not try to constrain the generation of patches, which has huge flaws in naturalness.

\section{Method}
\subsection{Motivation}
In the domain of general-purpose object detection, pre-trained generative models have been shown to be able to generate natural adversarial images, Hu \etal~\cite{hu2021naturalistic} with Biggan and Stylegan2, Diff-PGD~\cite{xue2024diffusion} with SDEdit. But When it comes to UAVs, there has been very little research on natural adversarial patches. We notice that the characteristics of UAV imagery make it difficult to generate adversarial patches. Firstly small adversarial patches lead to narrow down the solution space that can attack the object detector. Besides UAV imagery is faced with a more complex physical environment and has stricter requirements on patches. 

\subsection{Preliminaries}
\textbf{Threat Model}.
Given a training set (X, Y) and an object detector $f$, where X, Y and $f$ are the sampled images, ground truth labels and $f (\textbf{x}): \textbf{x} \rightarrow {\rm y}, \textbf{x}\in {\rm X},{\rm y} \in {\rm Y}$ that outputs the identified object y. Our adversarial goal is to find a patch $\textbf{P}$ that usually follows a square-sized setting where $\textbf{P}$ $\in$ $\mathbb{R}^{s\times s\times 3}$ and s accounts for the patch size, such that $f ((1-\textbf{m})\odot \textbf{x}+\textbf{m}\odot \textbf{P}) \neq {\rm y}$, where $\textbf{m}$ denotes a constructed binary mask that is 1 at the placement position of the adversarial patch and 0 at the remaining positions, $\odot$ denotes the Hadamard product (element product). In general, the EMA method is divided into three parts: Adversarial Perturbation, Text Guidance and Scene Matching. Define the entire process of generating adversarial patches as:
\begin{equation}
\textbf{P} = {\rm EMA}(\textbf{P}_{init},T,\gamma,\delta),
\end{equation}
where $P_{init}$ can sample from the background or Stable Diffusion, T means text description, $\gamma$ and $\delta$ are the input parameters of Stable Diffusion and scene matching. 
As a result, the generation of the adversarial patch is usually solved according to the following equation:
\begin{equation}
p^{*}(\textbf{x},{\rm y},\textbf{P},\textbf{m})={\rm argmax} \mathcal{L}(f ((1-\textbf{m})\odot \textbf{x}+\textbf{m}\odot \textbf{P}),{\rm y})
\end{equation}
\textbf{Diffusion Models}.We adopt a pre-trained Stable diffusion in text guidance. To better understand our work, it is useful to give an overview of Diffusion Models.
Denoising Diffusion Probabilistic Models(DDPM)~\cite{ho2020denoising} is a class of generative models that has gained significant attention in recent years for its ability to produce high-quality samples. 
DDPM consists of two main processes: the forward diffusion process and the denoising process.

The diffusion process is a Markov chain that gradually transforms data points (such as images) into noise. The diffusion process can be represented as:
\begin{equation}
    \textbf{x}_t = \sqrt{\alpha_t} \textbf{x}_{t-1} + \sqrt{1 - \alpha_t} \epsilon_t, \quad t = 1, 2, \ldots, {\rm T}
\end{equation}
where $x_t$ is the image at step $t$, $\alpha_t$ is the diffusion coefficient (which typically decreases with increasing $t$), $\epsilon_t$ is noise drawn from a standard normal distribution, and $T$ is the number of diffusion steps.

The denoising process is the reverse process of the diffusion process, aiming to recover the original data from the noise. In Stable Diffusion, the denoising process is usually implemented by a conditional neural network (such as U-Net) that predicts the original image based on the text description and the current noisy image. The denoising process can be represented as:
\begin{equation}
    \textbf{x}_{t-1} = \frac{1}{\sqrt{\alpha_t}} \left(\textbf{x}_t - \frac{1 - \alpha_t}{\sqrt{1 - \bar{\alpha}_t}} \epsilon_\theta(\textbf{x}_t, t)\right)
\end{equation}
where $\epsilon_\theta(\textbf{x}_t, t)$ is the noise predicted by the neural network, and $\bar{\alpha}_t = \prod_{s=1}^t \alpha_s$.

Stable Diffusion represents a family of latent diffusion
models~\cite{rombach2022high} which is capable of generating photorealistic images given any text input.
Stable Diffusion uses a text encoder to convert the text description into a fixed-length vector representation. the text vector is usually fused with the image features at multiple layers of the network. This can be done by simply concatenating the text vector with the image features or through more complex mechanisms (such as self-attention mechanisms).
Our approach is developed into the Stable Diffusion, which is capable of meeting the requirements for generating natural adversarial patches.

\subsection{EMA Method}
\label{4.2}
EMA method is born to solve two major problems that exist in adversarial attacks against UAV object detection: first, the adversarial patches generated by existing methods have poor concealment because they do not consider the influence of the physical environment; second, there is no good way to effectively restrict the search space and optimization area. To solve these two problems at the same time, stable diffusion is introduced into the EMA method, where the text guidance for color restriction is extremely important. To cope with the color changes that may be brought by the physical domain and to better integrate the adversarial patches into the environment in all aspects, we perform scene matching for the adversarial patches.
The following will provide a more detailed introduction to the method and its function.

\begin{figure}
    \centering
    \begin{subfigure}{.13\textwidth}
        \centering
        \includegraphics[width=\linewidth]{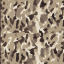}
    \end{subfigure}%
    \tikz[baseline] \draw[->, thick] (0.2,0.8) -- (1,0.8);
    \begin{subfigure}{.13\textwidth}
        \centering
        \includegraphics[width=\linewidth]{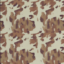}
    \end{subfigure}%
    \tikz[baseline] \draw[->, thick] (0.2,0.8) -- (1,0.8);
    \begin{subfigure}{.13\textwidth}
        \centering
        \includegraphics[width=\linewidth]{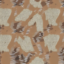}
    \end{subfigure}
    \tikz[baseline] \draw[->, thick] (0.2,0.8) -- (1,0.8);
    \begin{subfigure}{.13\textwidth}
        \centering
        \includegraphics[width=\linewidth]{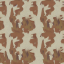}
    \end{subfigure}%
    \tikz[baseline] \draw[->, thick] (0.2,0.8) -- (1,0.8);
    \begin{subfigure}{.13\textwidth}
        \centering
        \includegraphics[width=\linewidth]{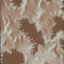}
    \end{subfigure}%
    \caption{Training process of the adversarial patch in EMA method.}
    \label{sand}
\end{figure}

\textbf{Text Guidance}.
Due to the characteristics of drones, the changes in pixels of adversarial patches will be more intense compared to general adversarial object detection. In our experiments, we found that the  GAN and SDEdit both failed to achieve good results and could not consider the impact of environmental colors in the absence of text guidance. Text, as a strong constraint condition, can well meet our requirements in adversarial drone target detection. Pre-trained Stable Diffusion can provide strong prior knowledge so that when we input prompts related to color, the search space can be naturally restricted. In our framework, the choice of prompt and guidance scale should be strict in color but lenient in texture. The adversarial perturbations added in the previous step are successfully hidden through stable diffusion and transformed into texture changes to attack the target detector. Without controlling the text, the model will gradually degrade into a method of directly optimizing the patch as the gradient is updated.
To better illustrate the entire training process, Figure.\ref{sand} lists the changes in the patch during the training process.

\textbf{Scene Matching}.
The images captured by UAVs are greatly affected by changes in the scene, so the generated adversarial patches need to consider physical conditions such as lighting, contrast, and noise. We define a set of physical conditions as $\phi$. To simulate the actual effect of the adversarial patch being placed on the roof of a car, it is necessary to scale the adversarial patch according to the size of the car and rotate it to simulate the change in the perspective of the UAV camera. The formula is as follows:
\begin{equation}
\textbf{I}_{atk} = {\rm A}(\textbf{I},\phi_{\delta}(\textbf{P}_{ema})),
\end{equation}
where $\textbf{I}$ is the image from the dataset, and A utilizes the victim object detector to recognize cars in the scene and renders the adversarial patch on the proper location by affine transformation.
\subsection{Overall Attack Process}
\begin{figure}[h]
\begin{center}
    \includegraphics[width=0.95\linewidth]{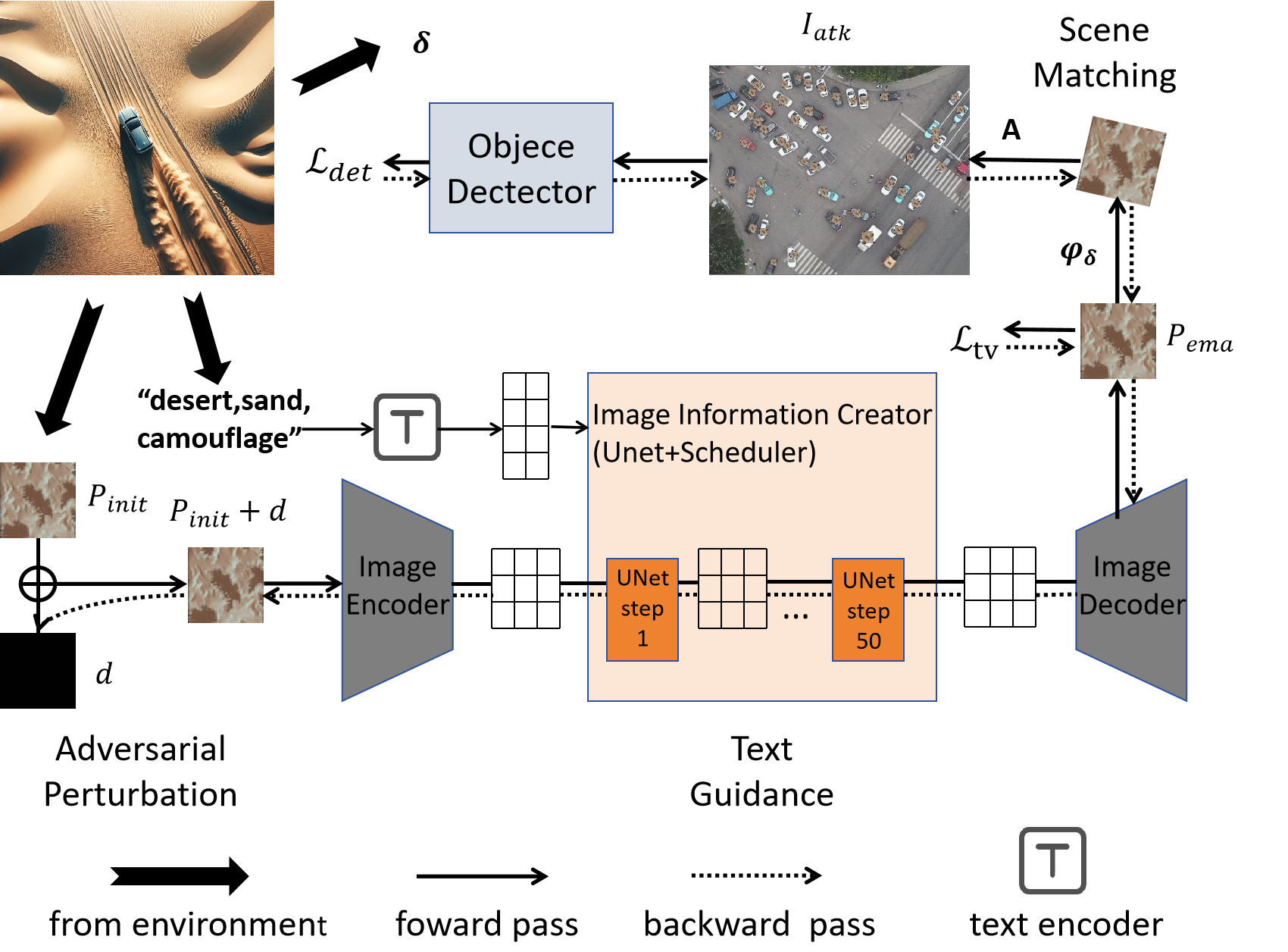}
\end{center}
    \caption{Overview of our Environmental Matching Attack framework which crafts the patches for object detectors by a pre-trained Stable Diffusion through the iterative optimization process for the adversarial perturbation patch.}
    \label{pipeline}
\end{figure}
In this section, we will provide an overall introduction to the EMA method framework and algorithm. Figure.\ref{pipeline} shows an overview of our framework including three parts. In addition to the text guidance and scene matching mentioned in Section \ref{4.2}, we will first focus on introducing the adversarial perturbation here.

Currently, the majority of adversarial patch methods directly optimize the adversarial patch itself, but this approach can cause significant changes to the original image to achieve good attack effects, which poses a great challenge to the naturalness of the adversarial patch. In contrast, since there are no hidden layers in the network, the model parameters can be set to a tensor $d$ with the same size as $P_{init}$ and a value of zero. Compared to directly optimizing the patch, adding adversarial perturbations has many advantages. Firstly, the perturbation can be seen as noise in the original image, which better matches the denoising process of stable diffusion, and expands the search range under the texture restriction on color, making it easier to find constrained optimal solutions. Secondly, this method involves fewer changes to the original image and it can preserve the information of the original image. From a macroscopic perspective, similar to PGD, it is like adding adversarial perturbations to $P_{init}$.

We exploit the $l_{\infty}$ norm to constrain d, and the formula for updating $P_{init}$ in each iteration is as follows:
\begin{equation}
\textbf{P}_{init} = {\rm Clip}(\textbf{P}_{init} + d).
\end{equation}
Clip is the clipping function defined in Eq.~\ref{eq6}.
\begin{equation}
{\rm Clip}(\textbf{P}) = \{p_{i}|p_{i}\leftarrow min(max(p_{i},\tau),0)\}
\label{eq6}
\end{equation}
where ${\rm p_{i}}$ is the i-th element of $\textbf{P}$.

After adding adversarial perturbations, the image carries adversarial information. Text guidance helps the diffusion model to pay more attention to the color information carried by the input image, achieving the effect of limiting the search range. Adversarial information is more likely to appear in the output image in the form of texture changes. The selection of the prompt should conform to the elements of the environment. As shown in Figure.\ref{pipeline}, the prompt we selected is "desert, sand, camouflage" with the first two words containing color information. Through our experiments, the word "camouflage" can expand the search range under color constraints, allowing the model to perform texture changes on the patches. The selection of text and guidance scale is crucial and requires careful consideration, as it can directly affect the training results of the model. 
\begin{algorithm}[t]  
  \caption{Patch Generation}  
  \label{Patch Generation}  
  \begin{algorithmic}[1]  
    \Require  
        Patch $\textbf{P}_{init}$, Prompt $T$, Time step $t$, Step size $s$, Adversarial perturbation $\textbf{d}$
     \Ensure $\textbf{P}_{ema}$
     \State $\textbf{x}_t=\textbf{P}_{init}+\textbf{d}$
     \Repeat
     \State $\textbf{x}_{t-s} = \sqrt{\alpha_{t-s}} \left( \frac{\textbf{x}_t - \sqrt{1 - \alpha_t} \cdot \epsilon_{\theta}(\textbf{x}_t,t,T)}{\sqrt{\alpha_t}} \right) + \sqrt{1 - \alpha_{t-s}} \cdot \epsilon_{\theta}(\textbf{x}_t,t,T)$
     \State t = t - s
     \Until t < 0
     \State $\textbf{P}_{ema} = \textbf{x}_{t+s}$
  \end{algorithmic}  
\end{algorithm}

With Algorithm.\ref{Patch Generation} and Scene Matching, $I_{atk}$ is generated to attack UAV object detector. Our patch optimization is implemented through the computation of two losses:
\begin{equation}
    \mathcal{L}_p = \lambda \mathcal{L}_{tv} + \mathcal{L}_{det}
\end{equation}
\begin{itemize}
\item[$\bullet$] 
Total variation loss is effective in removing noise while preserving edge information, resulting in smoother and clearer images. Compared to other smoothing techniques, total variation loss better preserves the edges and texture details of images, avoiding excessive blurring.
\begin{equation}
    \mathcal{L}_{tv} = \frac{\sqrt{\sum_{i}^{S} \sum_{j}^{S} \left(\textbf{P}_{i,j} - \textbf{P}_{i+1,j} \right)^2 + \left( \textbf{P}_{i,j} - \textbf{P}_{i,j+1} \right)^2}}{\rm N}
\end{equation}
where N denotes the number of pixels on the given adversarial patch $\textbf{P}$.
\item[$\bullet$] 
The detector loss is designed to reduce the accuracy of object detectors. The loss can be defined as follows:
\begin{equation}
\mathcal{L}_{\text{det}} = \frac{\sum_{i}^{\rm N} \text{conf}(\textbf{I}_{atk}) \times \text{obj}(\textbf{I}_{atk})}{\rm N}
\end{equation}
where conf and obj measure the object detector class confidence score and objectiveness score for the $\textbf{I}_{atk}$
\end{itemize}

\section{Experiments}
In this section, we first describe the implementation details of the proposed method and then conduct various qualitative and quantitative experiments on the proposed adversarial patches in both digital and physical domains to evaluate their attack effectiveness and naturalness and answer the following questions:

\begin{itemize}
	\item \textbf{Q1} : How does our method perform in balancing attacking performance and naturalness?
	\item \textbf{Q2} : Why is it difficult to balance attacking performance and naturalness in the domain of UAV?
	\item \textbf{Q3} : Why do we adopt stable diffusion instead of other generative models?
\end{itemize}
\begin{figure}[htb]
\subfloat{
  \begin{minipage}{0.23\textwidth}
    \centering
    \includegraphics[width=\linewidth]{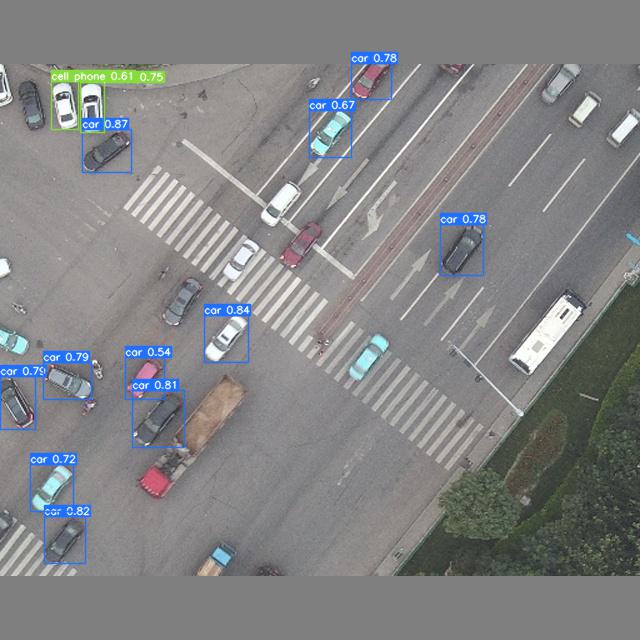}
    YOLOv5l\\DroneVehicle\\Patchless
  \end{minipage}
}
\subfloat{
  \begin{minipage}{0.23\textwidth}
    \centering
    \includegraphics[width=\linewidth]{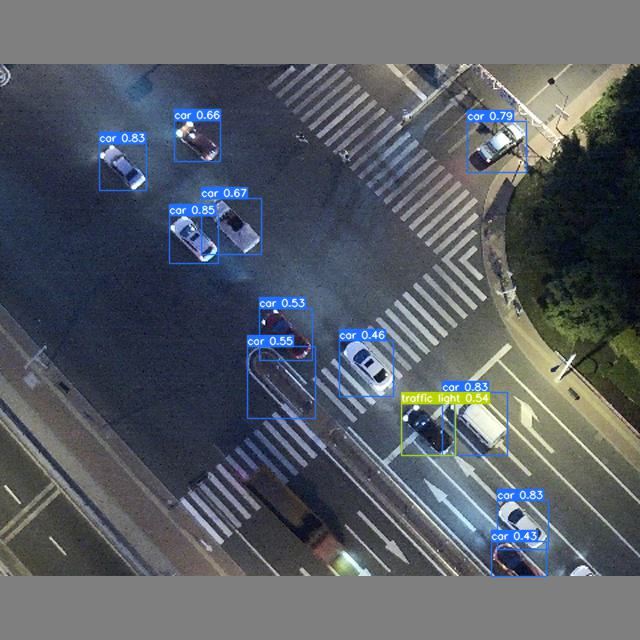}
    YOLOv5U\\DroneVehicle\\Patchless
  \end{minipage}
}
\subfloat{
  \begin{minipage}{0.23\textwidth}
    \centering
    \includegraphics[width=\linewidth]{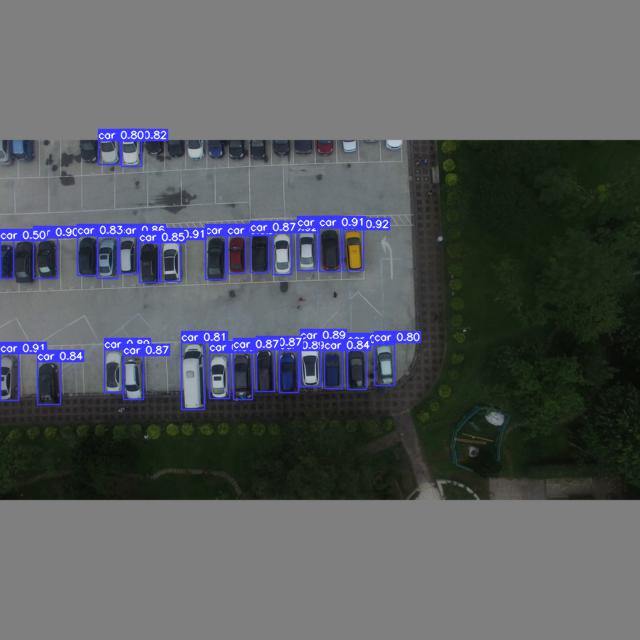}
    YOLOv5l\\Carpk\\Patchless
  \end{minipage}
}
\subfloat{
  \begin{minipage}{0.23\textwidth}
    \centering
    \includegraphics[width=\linewidth]{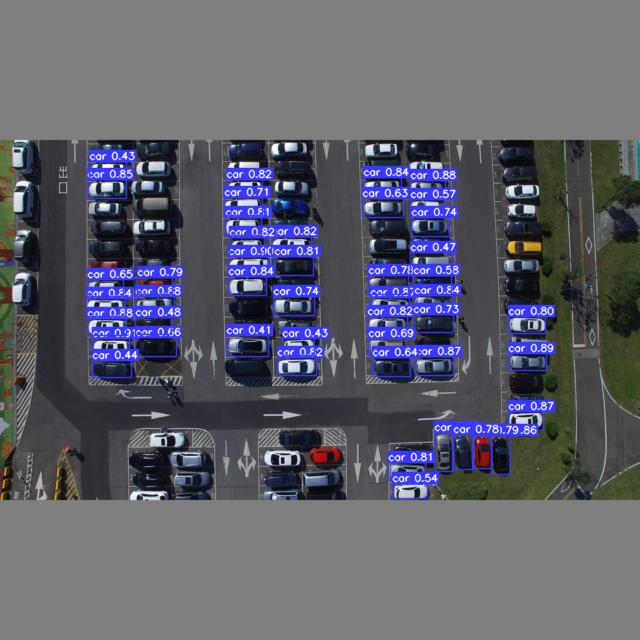}
    YOLOv5U\\Carpk\\Patchless
  \end{minipage}
}
\\
\subfloat{
  \begin{minipage}{0.23\textwidth}
    \centering
    \includegraphics[width=\linewidth]{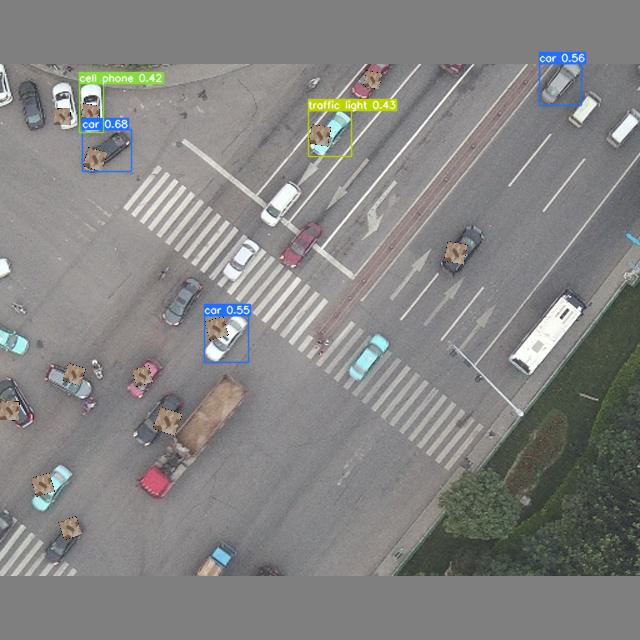}
    YOLOv5l\\DroneVehicle\\Patched
  \end{minipage}
}
\subfloat{
  \begin{minipage}{0.23\textwidth}
    \centering
    \includegraphics[width=\linewidth]{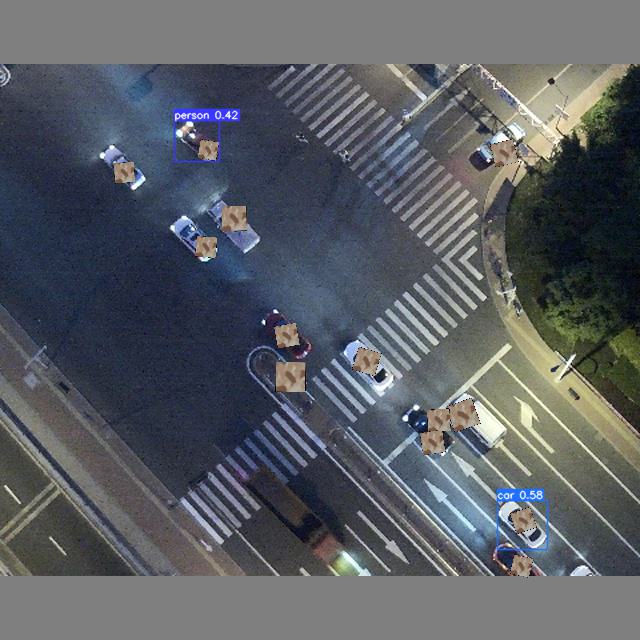}
    YOLOv5U\\DroneVehicle\\Patched
  \end{minipage}
}
\subfloat{
  \begin{minipage}{0.23\textwidth}
    \centering
    \includegraphics[width=\linewidth]{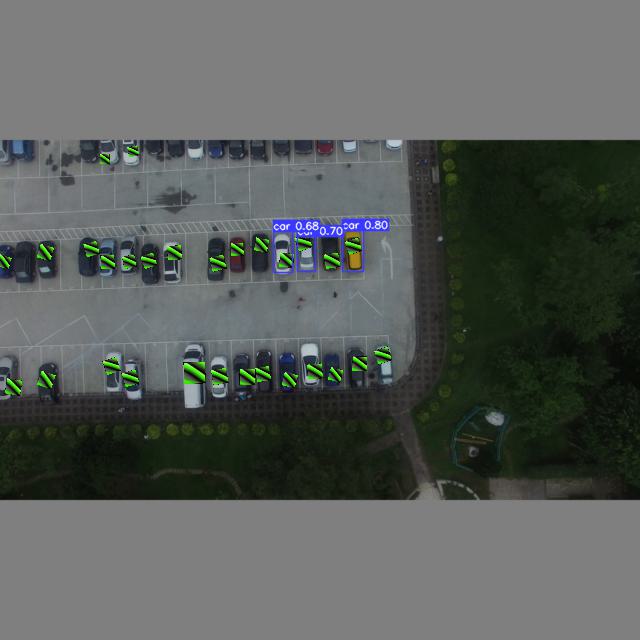}
    YOLOv5l\\Carpk\\Patched
  \end{minipage}
}
\subfloat{
  \begin{minipage}{0.23\textwidth}
    \centering
    \includegraphics[width=\linewidth]{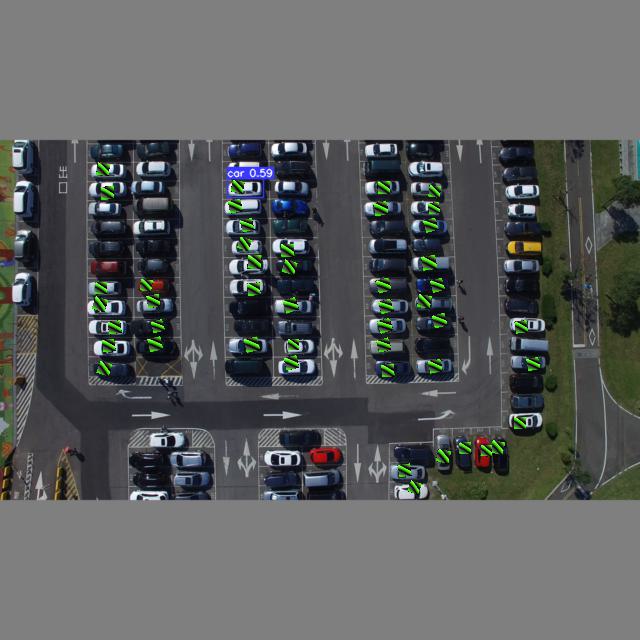}
    YOLOv5U\\Carpk\\Patched
  \end{minipage}
}
\caption{Samples on DroneVehicle and Carpk with and without adversarial patches.}
\label{digital}
\end{figure}
\subsection{Implementation details}
In our experiment, we use pre-trained stable diffusion, which is trained on 512x512 images from a subset of the LAION-5B dataset and uses a frozen CLIP ViT-L/14 text encoder to condition the model on text prompts. The prompt is often set to words related to color, which allows Stable Diffusion to limit color primarily rather than texture. For most of our results, we test the input parameters of Stable Diffusion and set the number of inference steps to 20, the guidance scale to 6.5, and others to default. In the process of training noise patch $d$, we control $d$ with $l_{\infty}$ norm, which is usually set to 0.6.
During the entire training process, Adam is used to optimize, where the learning rate is 0.005, the batchsize is 8, and the total epoch is 400. In addition, in the process of applying the adversarial patch to the UAV image, we randomly rotate it from -20° to 20° and scale it to 30$\%$ of the vehicle area.

\subsection{Datasets and victim models}
The experiment mainly uses two UAV benchmark datasets, DroneVehicle~\cite{sun2022drone} and Carpk~\cite{hsieh2017drone}.
The DroneVehicle dataset captures a total of 15,532 pairs (31,064) of images from day to night, including 441,642 labeled instances, with occlusion and scale changes in the real environment.
The Carpk dataset takes 1,448 images from a UAV at an altitude of about 40 meters, from nearly 90,000 cars in four different parking lots. 
To get closer to UAV object detection, two other UAV datasets Visdrone~\cite{cao2021visdrone} and UAVDT~\cite{du2018unmanned} are used to train yolov5~\cite{jocher2022ultralytics} to obtain yolov5V and yolov5U, together with the weights yolov5l and yolov5x from the well-known coco benchmark~\cite{lin2014microsoft} to form a set of victim models.

\subsection{Digital Attack Evaluations}
To answer these three questions, we first design a digital domain experiment that focuses on the attacking performance of adversarial patches in the cross-dataset and black-box situations that are common in actual UAV adversarial attacks. 
We implement the proposed method to generate environmental matching adversarial patches on the Carpk dataset and test them on the DroneVehicle dataset. Fig.\ref{digital} shows the attacking performance of adversarial patches of the EMA method in the digital attack. The attack examples are adversarial patches with sand and grass colors as the main colors respectively.

The criteria for evaluating attacking performance and naturalness are as follows. According to the method proposed in~\cite{hu2021naturalistic}, we calculate the Mean Average Precision (mAP) considering the Intersection over Union (IoU) threshold from 0.5 to 0.95. In this case, we normalized the detector's performance on clean images to a mAP of 100$\%$ for direct comparison of performance degradation. 
In consideration of environmental matching naturalness, we define it as the difference between the main color of the adversarial patch and the color of the environment (the main color of the initial patch), measured by the CIE2000 score ($\Delta E_{00}$) which is regarded as the best uniform color-difference model coinciding with subjective visual perception. 
To answer \textbf{Q1}, we compare our method, the baseline, and the adjusted baseline, and Table \ref{CIEDE2000} clearly shows that we achieved a good trade-off between attack performance and naturalness.
Since the existing methods did not consider naturalness in the environment, we add the main color of the environment to the adversarial patches generated by the baseline method for a better comparison.
\begin{table}[htb]
\begin{center}
\begin{threeparttable}
\setlength{\tabcolsep}{2.2mm}
  \caption{ Evaluations in mAP(\%) and $\Delta E_{00}$.}
  \label{CIEDE2000}
  \begin{tabular}{@{}lllll|l@{}}
    \toprule
    Method & YOLOv5l & YOLOv5x & YOLOv5V & YOLOv5U &$\Delta E_{00}\downarrow$\\
    \midrule
${ }^{(P 1)}$Shrestha \etal.~\cite{shrestha2023towards} & 6.5 & 3.6 & 5.2 & 12.1&58.18
\\
${ }^{(P 2)}$Shrestha \etal.* & 11.0 & 5.8 & 8.4 & 15.2&15.23 \\
${ }^{(P 3)}$Ours-YOLOv5V & 6.8 & 5.0 & 6.8 & 12.0 &13.12\\
\midrule

${ }^{(P 4)}$Thys \etal.~\cite{thys2019fooling} & 4.9 & 3.9 & 14.5 & 20.2 &56.05
\\
${ }^{(P 5)}$Thys \etal.* & 9.9 & 7.6 & 16.4 & 24.2 &7.10 \\
${ }^{(P 6)}$Ours-YOLOv5l & 5.5 & 5.1 & 15.5 & 21.4& 6.60 \\
\bottomrule
\end{tabular}
\begin{tablenotes}
\item *modified for comparison
\end{tablenotes}
\begin{figure}[H]
\vspace{-2em}
\begin{center}
\subfloat[P1]{
    \includegraphics[scale=0.6]{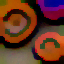}}
\subfloat[P2]{
    \includegraphics[scale=0.6]{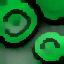}}
\subfloat[P3]{
    \includegraphics[scale=0.6]{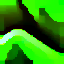}}
\subfloat[P4]{
    \includegraphics[scale=0.6]{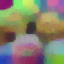}}
\subfloat[P5]{
    \includegraphics[scale=0.6]{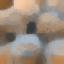}}
\subfloat[P6]{
    \includegraphics[scale=0.6]{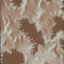}}
    \end{center}
\end{figure}
  \end{threeparttable}
  \end{center}
  \vspace{-6em}
\end{table}

It can be observed that although the existing methods have excellent attacking performance, which is sensitive to the color of the adversarial patch. Therefore, we add the main color of
 the environment to the patch from the baseline so that the color difference between them is close.
This is reflected in the values of the $\Delta E_{00}$.
It is obvious that when we try to increase the naturalness of the baseline, its attacking performance significantly decreases. To demonstrate that our EMA method can adapt to various environments, we generate adversarial patches that can be placed in desert and grassland environments, primarily in sandy yellow and green colors, respectively. Table \ref{map} shows patches attacking performance on different victim object detectors. We mainly consider two scenarios: desert and grassland. In the white-box settings of yolov5l and yolov5x, we obtain the initial image and the Scene Matching parameters based on the environment pictures of the desert. 
\begin{table}[htb]
\vspace{-2.2em}
\begin{center}
\begin{threeparttable}
\setlength{\tabcolsep}{2.3mm}
  \caption{Different patches evaluations in mAP(\%) for the DroneVehicle dataset.
  }
  \label{map}
  \centering
  \begin{tabular}{@{}lllll@{}}
    \toprule
    \begin{tabular}{ll}
\diagbox{Trained on}{Victim} &  \\
\end{tabular} & YOLOv5l & YOLOv5x & YOLOv5V & YOLOv5U  \\
    \midrule
${ }^{(P 1)}$ Ours-YOLOv5l & \textbf{5.5} & 5.1 & 15.5 & 21.4  \\

${ }^{(P 2)}$ Ours-YOLOv5x & 6.4 & \textbf{3.9} & 16.2 & 21.8  \\

${ }^{(P 3)}$ Ours-YOLOv5V & 6.8 & 5.0 & \textbf{6.8} & 12.0 \\

${ }^{(P 4)}$ Ours-YOLOv5U & 9.0 & 4.5 & 14.4 & \textbf{11.6} \\
\midrule
${ }^{(P 5)}$ Thys \etal.~\cite{thys2019fooling} & 4.9 & 3.9 & 14.5 & 20.2 
\\
${ }^{(P 6)}$ Thys \etal.* & 9.9 & 7.6 & 16.4 & 24.2 \\
${ }^{(P 7)}$ Shrestha \etal.~\cite{shrestha2023towards} & 6.5 & 3.6 & 5.2 & 12.1
\\
${ }^{(P 8)}$ Shrestha \etal.* & 11.0 & 5.8 & 8.4 & 15.2 \\
\midrule
${ }^{(P 9)}$ Hu \etal.~\cite{hu2021naturalistic} &5.7  & 2.9 & 4.1 & 11.1 \\
${ }^{(P 10)}$ Hu et al.* & 8.9 & 6.1 & 16.0 & 15.9  \\

${ }^{(P 11)}$ Diff-PGD~\cite{xue2024diffusion} & 7.7 & 2.6 & 5.0 & 6.7 \\
${ }^{(P 12)}$ Diff-PGD* & 8.5 & 3.7 & 14.3 & 18.3  \\

\midrule\midrule
${ }^{(P 13)}$ Random & 11.4 & 11.6 & 19.0 & 26.3\\ 
\bottomrule
\end{tabular}
\begin{tablenotes}
\item *modified for comparison
\end{tablenotes}
\begin{figure}[H]
\vspace{-2em}
\begin{center}
\subfloat[P1]{
    \includegraphics[scale=0.6]{Figure/p1.png}}
\subfloat[P2]{
    \includegraphics[scale=0.6]{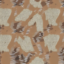}}
\subfloat[P3]{
    \includegraphics[scale=0.6]{Figure/p3.png}}
\subfloat[P4]{
    \includegraphics[scale=0.6]{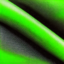}}
\subfloat[P5]{
    \includegraphics[scale=0.6]{Figure/p5.png}}
\subfloat[P6]{
    \includegraphics[scale=0.6]{Figure/p6.png}}
\subfloat[P7]{
    \includegraphics[scale=0.6]{Figure/p7.png}}
\subfloat[P8]{
    \includegraphics[scale=0.6]{Figure/p8.png}}
\end{center}
\vspace{-1em}
\begin{center}
\subfloat[P9]{
    \includegraphics[scale=0.6]{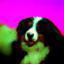}}
\subfloat[P10]{
    \includegraphics[scale=0.6]{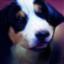}}
\subfloat[P11]{
    \includegraphics[scale=0.6]{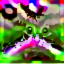}}
\subfloat[P12]{
    \includegraphics[scale=0.6]{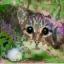}}
\subfloat[P13]{
    \includegraphics[scale=0.6]{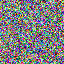}}
\end{center}
\end{figure}
  \end{threeparttable}
  \end{center}
\vspace{-5em}
\end{table}

We set the prompt to "desert, sand, camouflage" and obtain P1 and P2 after training. In the white-box settings of yolov5V and yolov5U, we consider generating adversarial images that match the primary color of the grassland, as the weights of these two object detectors were trained on a UAV benchmark dataset, representing stronger capabilities. Accordingly, we appropriately relax the selection of prompt and guidance scale, adding "black" to the prompt without affecting naturalness.

The results of the third part of Table \ref{map} effectively answer \textbf{Q2} and \textbf{Q3}. When we transfer the general-purpose natural adversarial attacks to the UAV domain, it is challenging for both GANs and image-to-image diffusion models like SDEdit to generate natural objects, even though they perform well in the general adversarial domain. P9 is trained on under the condition of $l_{\infty} = 50$, while P10 is trained with $l_{\infty} = 10$. 
As you can see, the purple background appeared unexpectedly, which looks very unnatural and catches our eyes in the desert or grass background.
Similarly, the same situation applies to P11 and P12.
At this point, the textual guidance of the text-to-image model becomes important. Although achieving good attacking performance with a photo of a dog pasted on a car is still difficult, our experiments show that it is possible to roughly control the color to match the environment and enhance the naturalness of the attack while maintaining a good level of attacking performance.

\subsection{Physical Attack Simulation}
To assess the effectiveness of the generated adversarial patches in real-world scenarios, we simulate the actual UAV environment using a car model and make a 2cm x 2cm patch that could perfectly cover the roof of the car model. To simulate the drone's altitude of 30-60 meters, We use Canon EOSR5 to photograph the patched and unpatched model cars in sand, grass, and other environment. We selected a YOLOv5U-trained patch and utilized YOLOv5U as the evaluation detector. We quantified the physical attack performance based on mAP. Table \ref{phy1} shows the attacking performance of our adversarial patches designed for different environments under physical conditions. We focused on testing the performance on grasslands and surprisingly found that, although there is a certain gap with the baseline in the digital domain, thanks to the consistency with the environmental color, the mAP in the physical domain is \textbf{27.4}, which is better than the 30.8 of the baseline method.

\begin{table}[H]
\caption{Attacking performance in different places.}
\label{phy1}
\begin{tabular}{ccccc}
\hline
Image
&
\includegraphics[width=0.20\linewidth]{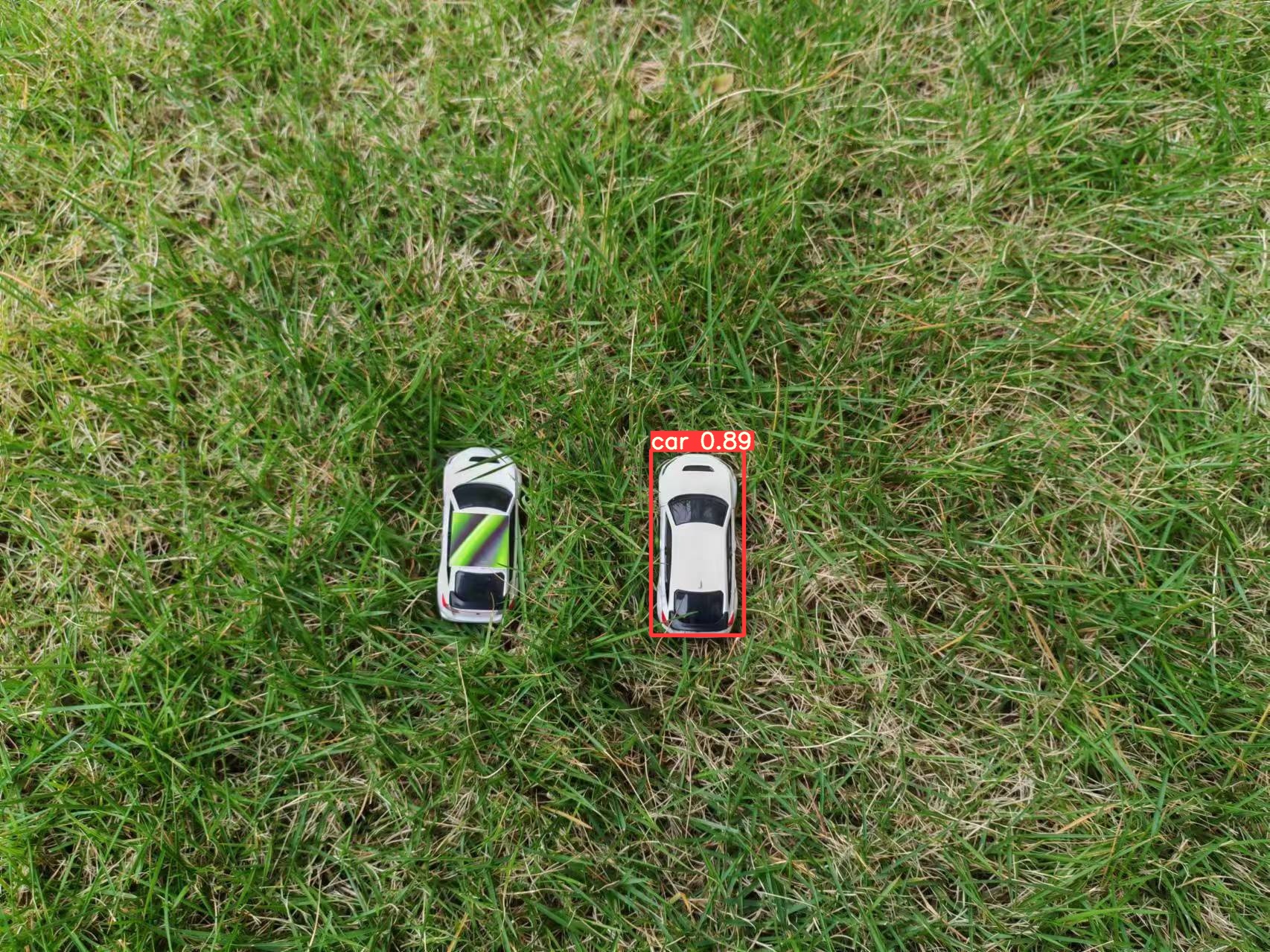} & 
\includegraphics[width=0.20\linewidth]{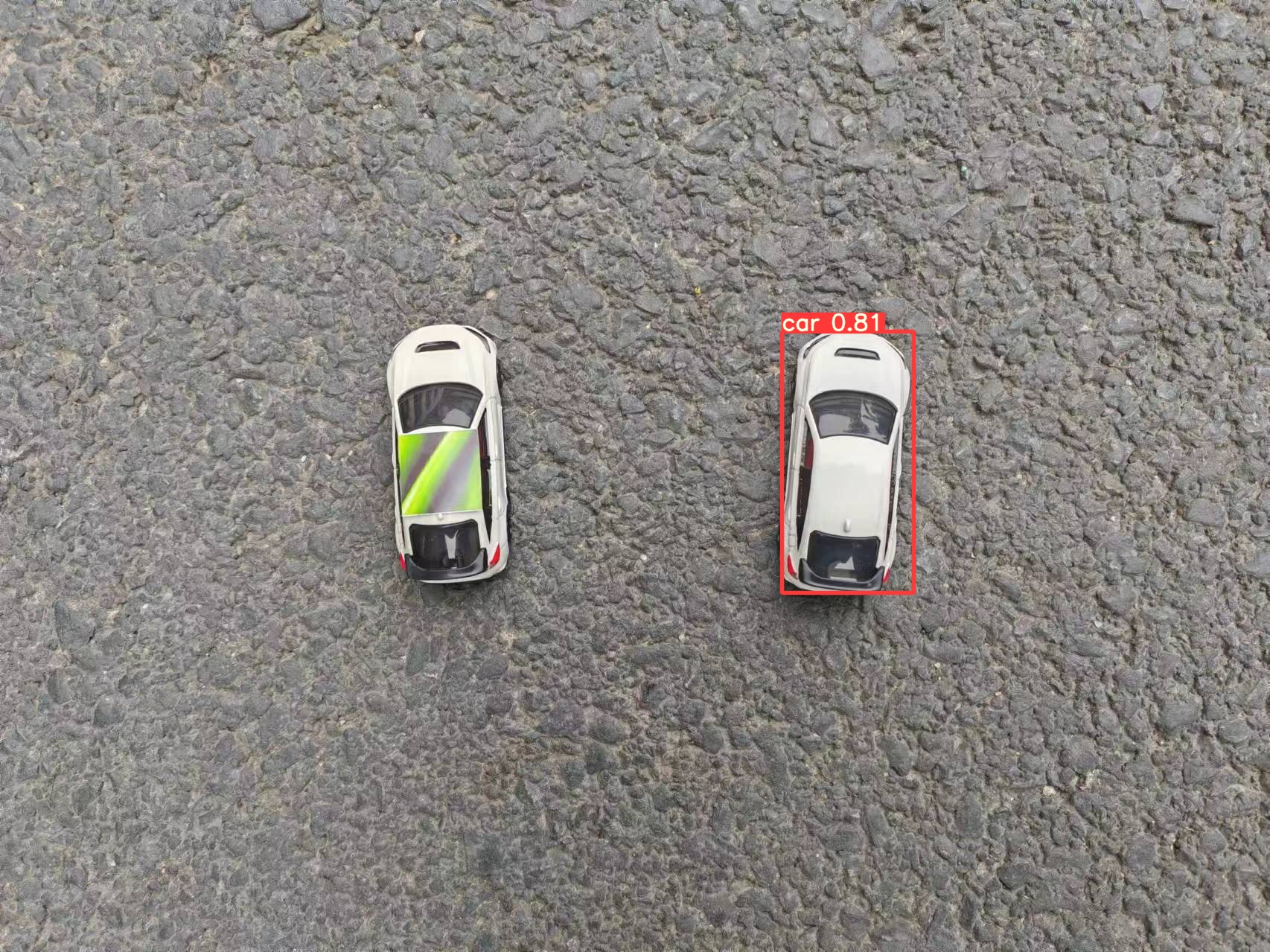} & 
\includegraphics[width=0.20\linewidth]{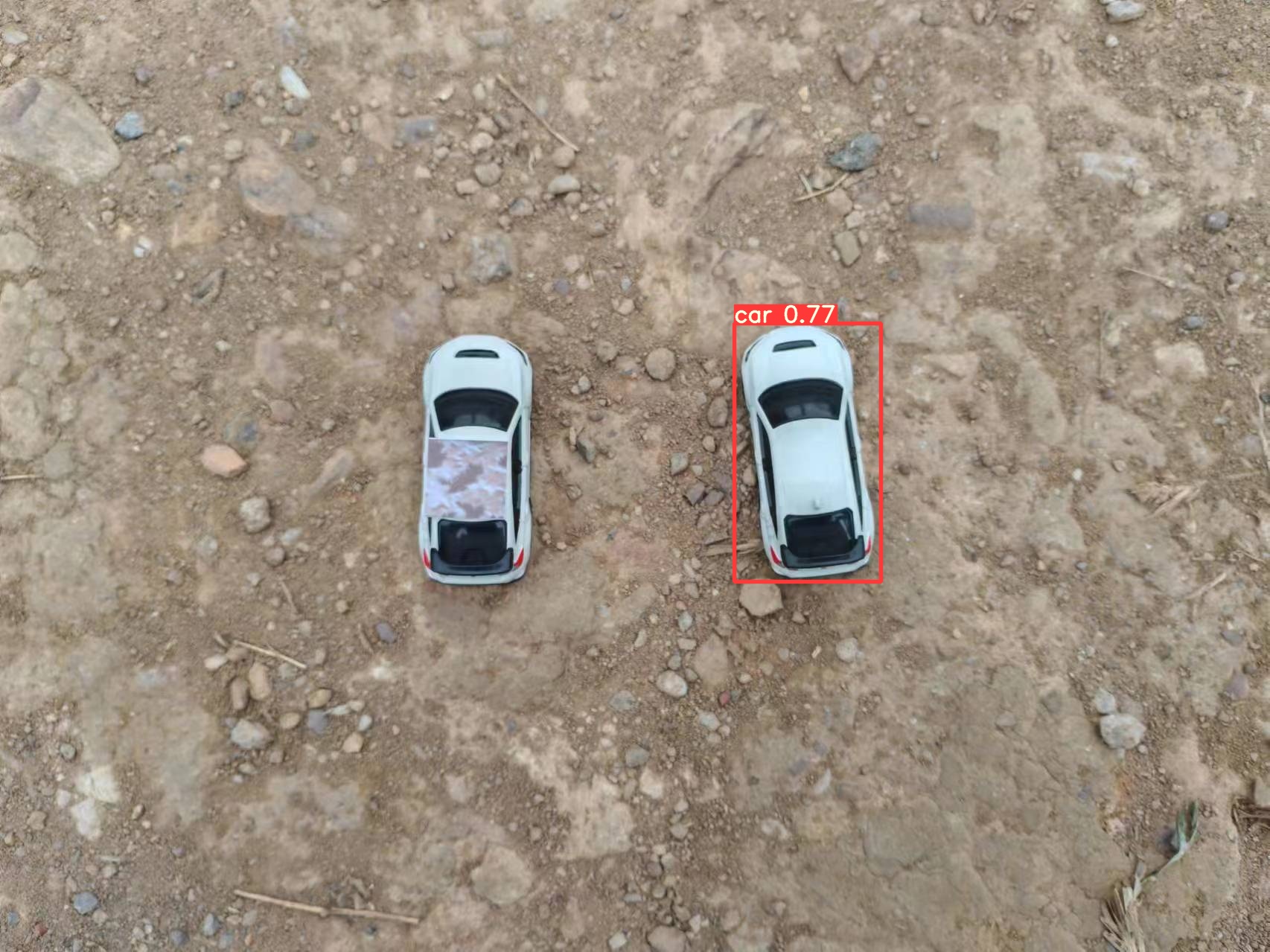} & 
\includegraphics[width=0.20\linewidth]{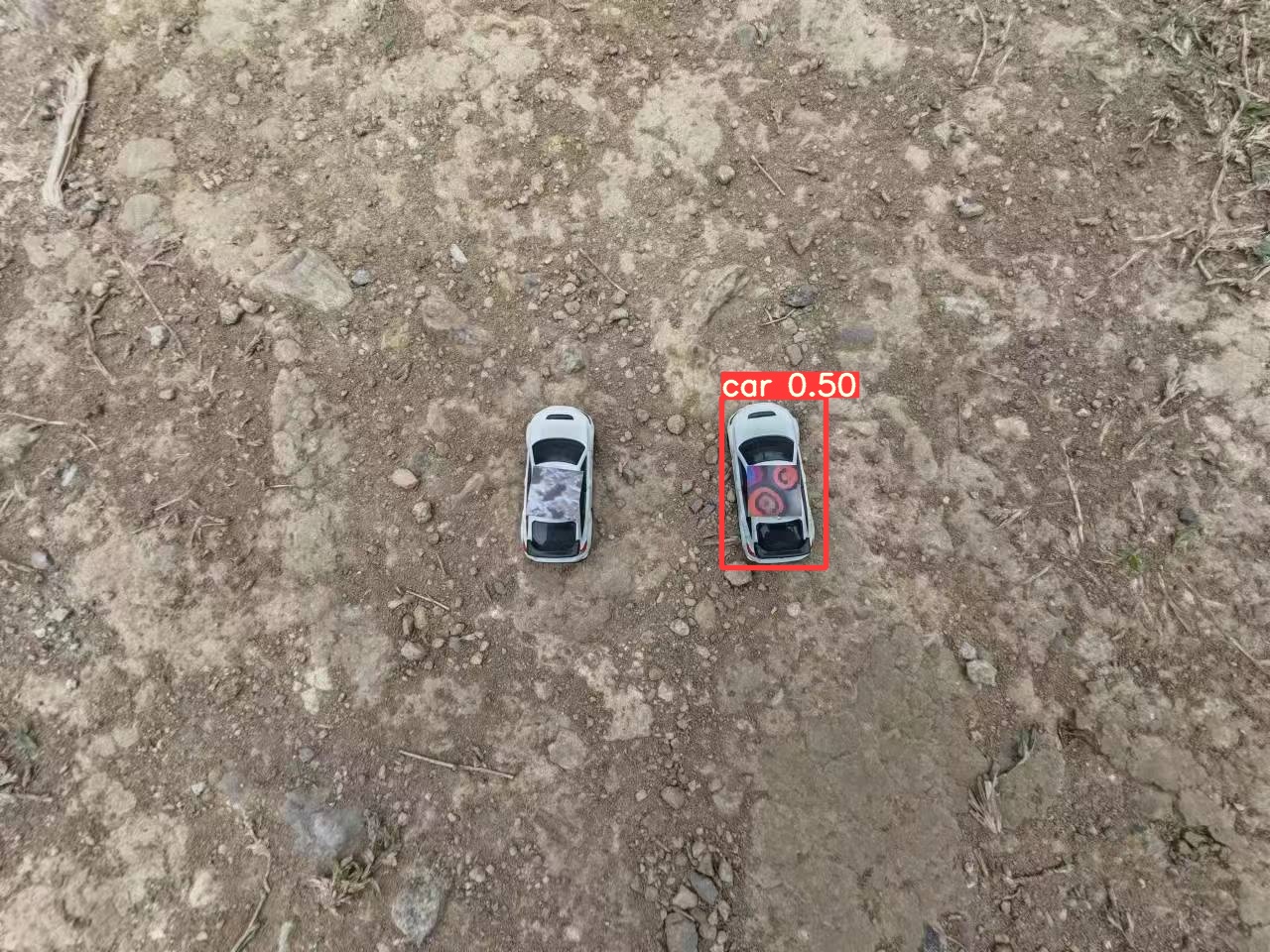} \\
\hline

Location & grass & clean & sand & sand \\
Comparison & patchless & patchless & patchless & Shrestha \etal \\
\hline
\end{tabular}
\end{table}

\subsection{Visualization}
\begin{figure}[H]
\centering
\subfloat[P4(our) and clean]{
		\includegraphics[scale=0.048]{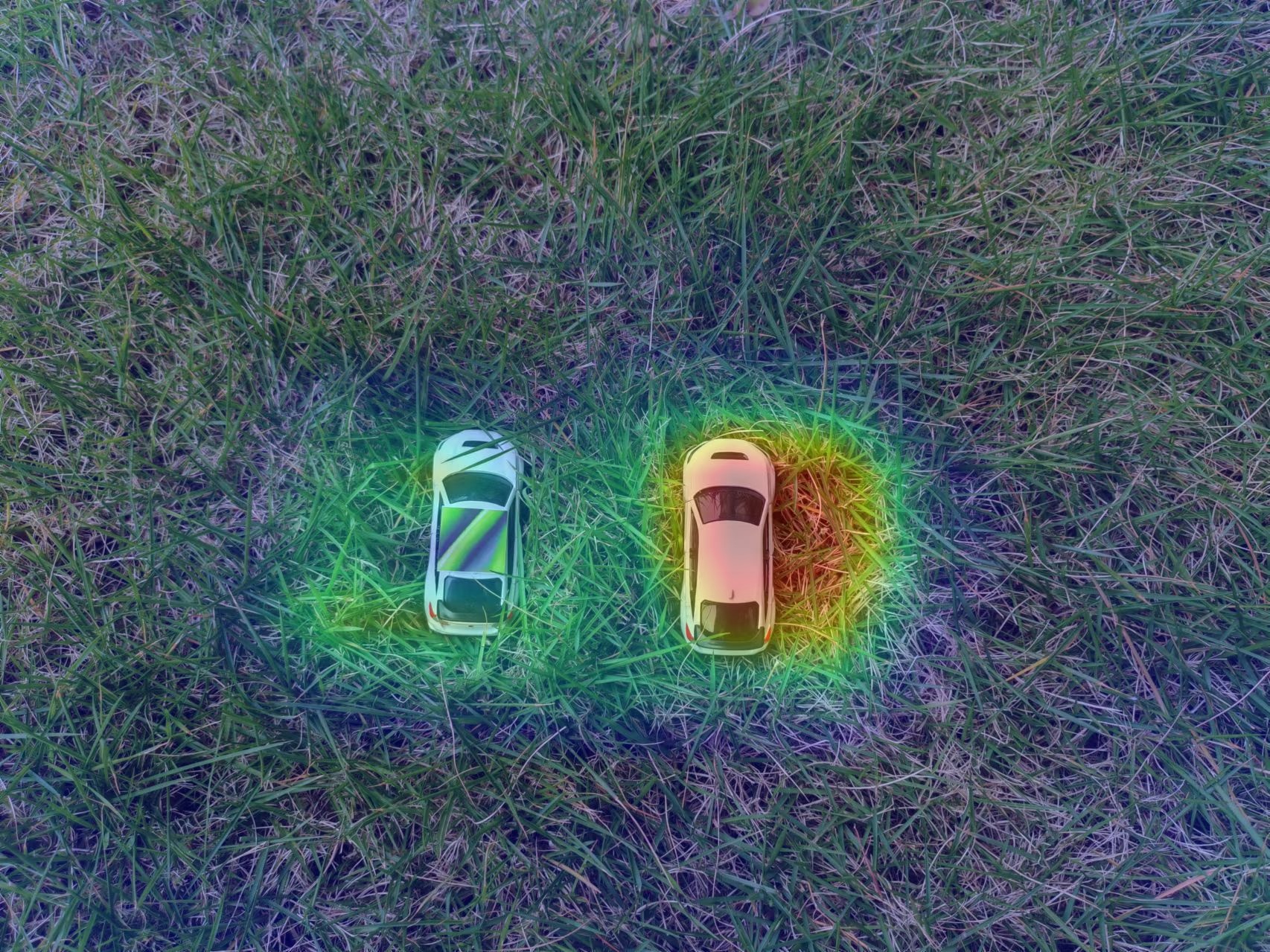}}
\subfloat[P9, P4(our) and P19]{
		\includegraphics[scale=0.048]{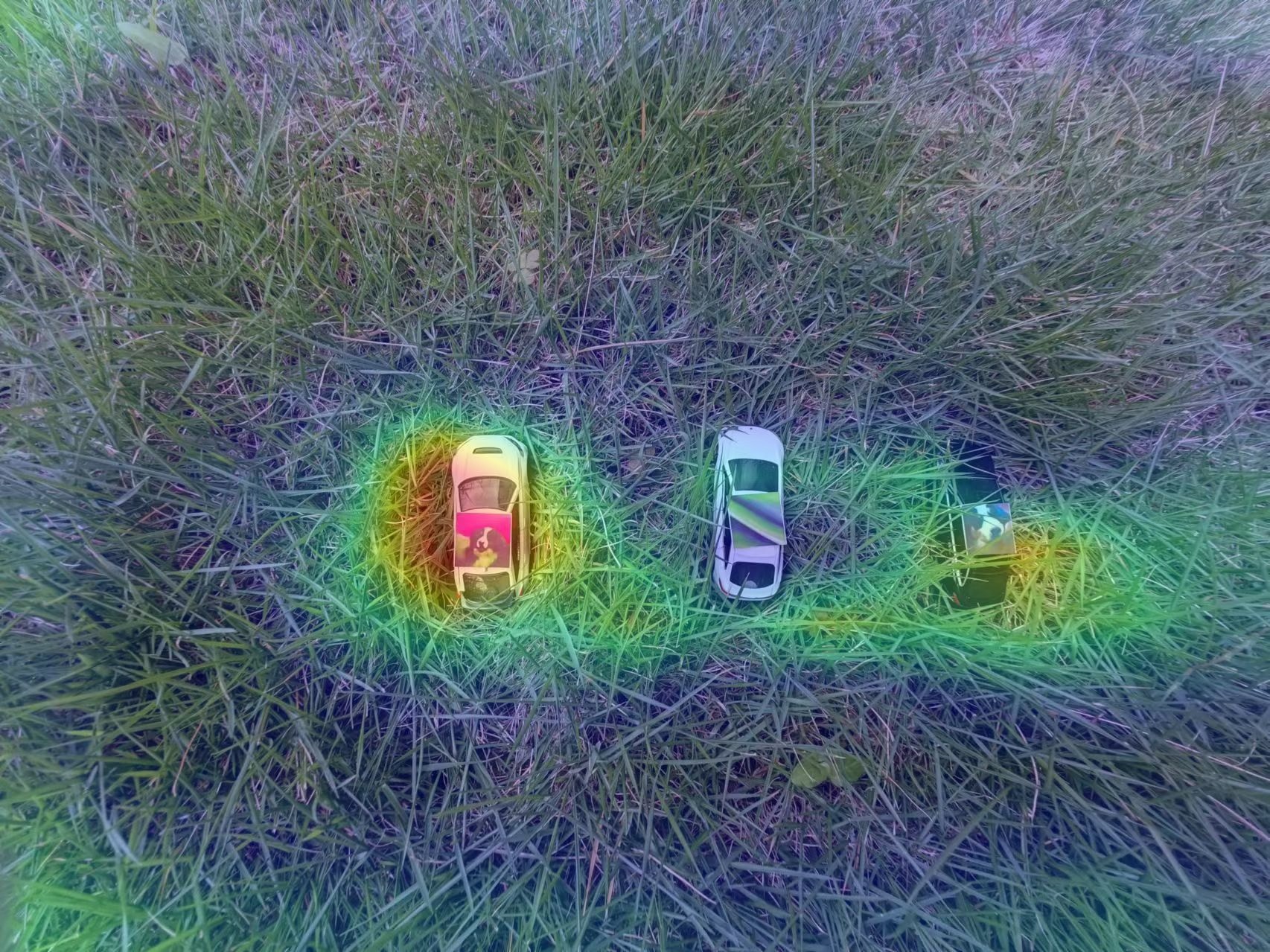}}
\subfloat[P1(our) and P10]{
		\includegraphics[scale=0.064]{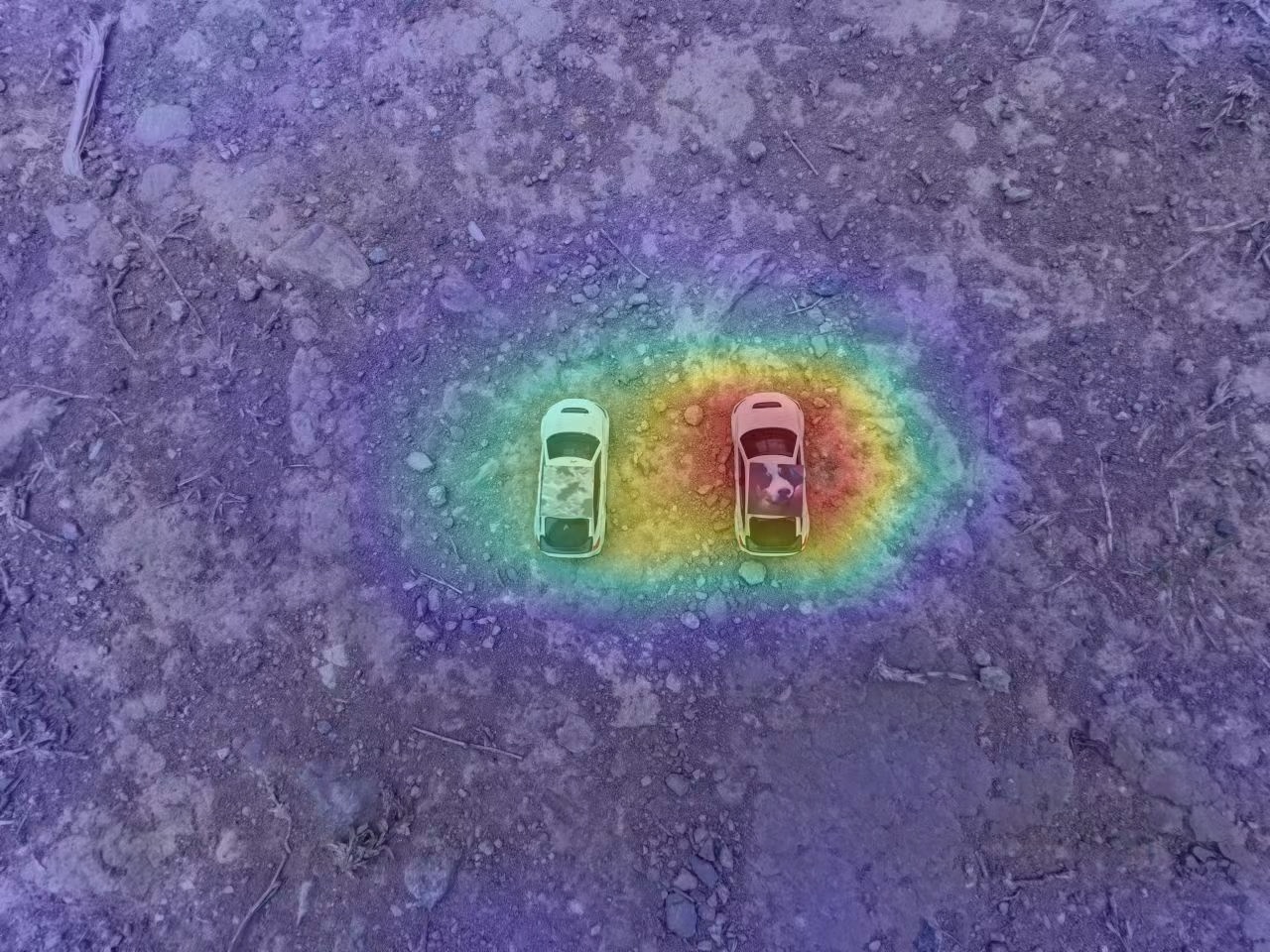}}
\subfloat[P1(our) and P7]{
		\includegraphics[scale=0.064]{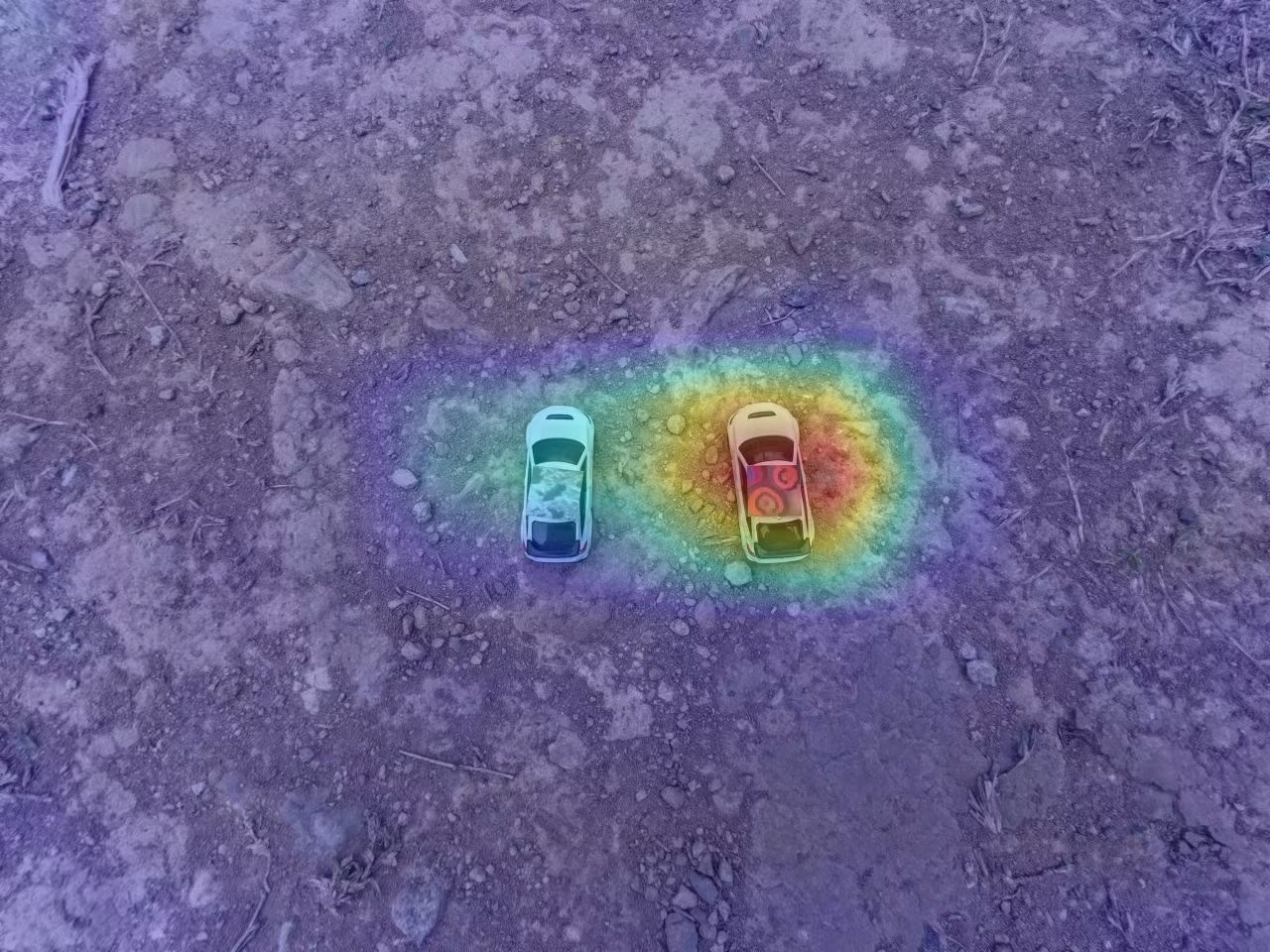}}
\caption{Comparerism with baseline in the visualization of human attention.}
\label{cam}
\end{figure}

Grad-CAM~\cite{selvaraju2017grad} is a method used to explain deep learning model predictions. It generates a heat map by weighting the feature maps of a specific layer by the predicted class probabilities of the model.
Li \etal~\cite{li2023towards} apply Grad-CAM to simulate human attention to get Mean Opinion Score(MOS). We use ResNet50~\cite{he2016deep} that pretrained on ImageNet to extract the attention map as a visualization of human attention. Compared with other adversarial patches, our EMA patches in specific scenarios are not easily noticed by humans. Results are illustrated in Figure.\ref{cam}

\section{Conclusion}

The widespread application of object detection in the UAV domain is easy to be attacked with adversarial patches. However,  existing adversarial patches struggle to strike a balance between naturalness and attack success. We propose new requirements for the naturalness of adversarial patches which is environmental naturalness and are the first to research generation methods for natural adversarial patches in the UAV domain. Leveraging the powerful prior knowledge provided by pre-trained stable diffusion, we introduce the Environmental Matching Attack, which can constrain the color of adversarial patches based on the environment, achieving environmental naturalness.  We also model the entire framework as a process of adding adversarial noise and denoising within a limited color space, achieving a good balance between attacking performance and naturalness. Extensive experiments have shown that we can achieve the naturalness of patches with an attack success rate that is comparable to the SoTA approach.

\clearpage


%
%
\bibliographystyle{splncs04}
\bibliography{main}
\end{document}